\documentclass[sigconf]{acmart}
\pdfoutput=1
\def\preabstract{}
\def\postabstract{}

\setcopyright{none}
\renewcommand\footnotetextcopyrightpermission[1]{} 
\title[Evolving Boolean Functions via Genetic Programming]{Evolving Boolean Functions with Conjunctions and Disjunctions via Genetic Programming}
\author{Benjamin Doerr}
\affiliation{\raisebox{0mm}[0mm][0mm]{\'E}cole Polytechnique, CNRS \\
Laboratoire d'Informatique (LIX) \\ Palaiseau, France}
\author{Andrei Lissovoi}
\affiliation{Deparment of Computer Science \\ University of Sheffield \\ Sheffield, United Kingdom}
\author{Pietro S. Oliveto}
\affiliation{Deparment of Computer Science \\ University of Sheffield \\ Sheffield, United Kingdom}



\let\postabstract\maketitle
\newcommand{\makebibliography}[1]{\bibliographystyle{ACM-Reference-Format}\bibliography{#1}}

\newcommand{\proofOf}{Proof of\xspace}

\newcommand{\showDOI}[1]{\unskip}
\newcommand{\showURL}[1]{\unskip}

\usepackage{amsmath,amssymb,xspace}
\usepackage{amsthm}
\usepackage[noend]{algpseudocode}
\usepackage{algorithm}
\usepackage{xcolor}
\usepackage{hyperref}
\newcommand{\andn}{\textsc{AND}\ensuremath{_n}\xspace}
\newcommand{\RLSGP}{RLS-GP\xspace}

\newcommand{\hh}{\ensuremath{\hat{h}}\xspace}
\newcommand{\Tmax}{\ell}
\newcommand{\N}{\mathbb{N}}

\usepackage{tikz}
\usetikzlibrary{trees}
\usepackage{pgfplots}
\pgfplotsset{compat = 1.14}

\begin{document}

\preabstract

\begin{abstract}
Recently it has been proved that simple GP systems can efficiently evolve the conjunction of $n$ variables if they are equipped with the minimal required components.
In
this paper, we make a considerable step forward by analysing the behaviour and performance of the GP system for evolving a Boolean function with unknown components, i.e., the function may consist of both conjunctions and disjunctions.
We rigorously prove that if the target function is the conjunction of $n$ variables, then the RLS-GP
using the complete truth table to evaluate program quality
evolves the exact target function in $O(\Tmax n \log^2 n)$ iterations in
expectation, where $\Tmax \geq n$ is a limit on the size of any accepted tree. When, as in realistic applications,
only a polynomial sample of possible inputs is used to evaluate solution
quality, we show how RLS-GP can evolve a conjunction with any polynomially small generalisation error with probability $1 - O(\log^2(n)/n)$.
To produce our results we introduce a
super-multiplicative drift 
theorem that gives significantly stronger runtime bounds when the expected progress is only slightly super-linear in the distance from the optimum. 
\end{abstract}

\postabstract

\section{Introduction}
Genetic Programming (GP) uses principles of Darwinian evolution to
evolve computer programs with some desired functionality. The most popular and
well-known GP approach, pioneered by Koza \cite{Koza92book}, represents
programs using syntax trees. It uses genetic algorithm inspired variation operators 
to search through the space of programs that may be generated with the available components, and principles of natural selection to favour
the ones which exhibit better behaviour on a wide variety of possible
inputs. In this setting, program quality is evaluated by executing the
constructed programs and comparing their output to the desired one.

Despite the many examples of successful applications of GP (see e.g. \cite{Koza10,LiuS13,BartoliCLM14}), there is only
a limited rigorous understanding of its behaviour and performance.
While theoretical analyses exist, the available results have considered
simplified GP systems, namely the \RLSGP and (1+1)~GP algorithms, 
which evolve a program by applying a simple tree-based mutation operator called HVL-Prime to a single individual at a
time 
for the evolution of non-executable tree structures \cite{DurrettNO11,DoerrKLL17,KotzingSNO14}, 
hence with no input/output behaviours.

Only recently, it has been proven that Boolean
conjunctions of $n$ variables can be evolved by \RLSGP~\cite{MambriniO16} and
(1+1)~GP~\cite{LissovoiO18} algorithms in an expected polynomial number of iterations. The evolved conjunctions are exact when
the complete truth table (i.e., the set of all $2^n$ possible inputs) is used to
evaluate solution quality, and generalise well when fitness is, more realistically, evaluated by sampling a
polynomial number of inputs uniformly at random from the complete truth table
in each iteration (i.e. employing Dynamic Subset Selection \cite{GathercoleR94}
to limit the total computational effort required to be polynomial with respect
to the problem size).

While the mentioned results are promising, the considered GP systems were considerably different to those used in practice. In particular, they  
were required to evolve a simple arity-$n$
Boolean conjunction from only its basic components (i.e. only the \textsc{AND} binary
Boolean operator, and the inputs necessary for the problem).
However in realistic applications, GP systems have access to a wider range of components than strictly necessary, because the required set of components
is not necessarily known in advance. 
Ideally, the system should be equipped with a complete set of operators (i.e., a set from which any Boolean function may be constructed).

In this paper, we make a considerable step forward by analysing the behaviour and performance
of \RLSGP for evolving an unknown Boolean function. More precisely, while the target function we consider is still \andn,
the conjunction of $n$ variables, the GP system has access to both
the binary conjunction (i.e., \textsc{AND}) and disjunction operators (i.e., \textsc{OR}).
Using \andn as the target function simplifies our understanding of the quality of candidate solutions that mix conjunction and disjunction operators. 

This more complex problem setting induces us to introduce more sophisticated features into the 
\RLSGP system than those necessary to evolve conjunctions using the \textsc{AND} operator alone, thus making the GP system more similar to realistic applications.
Since the presence of disjunctions in the current
solution may reduce the effectiveness of the mutation operator at producing
programs with better behaviour, 
we introduce a limit on the size of the syntax tree. This allows us to avoid issues due
to bloat (a common problem for GP systems, where the size of the solution is
allowed to increase without a corresponding increase in solution quality
\cite{Koza92book,PoliLM2008}). While alternative bloat control measures, such
as lexicographic parsimony pressure \cite{LukeP02}, would
prevent \RLSGP from adding any unnecessary disjunctions entirely, a
limit on the tree size is likely required to avoid pathological cases for more sophisticated insertion 
operators such as that of the (1+1)~GP, which would be able to accept disjunctions if the mutation operator simultaneously improves the solution in some other fashion.

With the limit on the tree size in place, our theoretical analysis reveals that the HVL-Prime mutation operator used in previous work~\cite{DurrettNO11,LissovoiO18}, which either inserts, substitutes or deletes one node of the tree,
may get stuck on local optima. Hence, the expected runtime of \RLSGP with the traditional  HVL-Prime operator has infinite expected runtime. To this end we introduce a mutation mechanism closer to the most commonly used subtree mutation \cite{Koza92book,PoliLM2008}, specifically allowing deletion to remove entire subtrees in one operation, rather than limiting it to only a single leaf and its immediate parent.

We show that \RLSGP with the above modifications is able to cope efficiently with the extended function set. In particular we prove that using the complete truth table to evaluate program quality, and rejecting any tree with more than $\Tmax = (1+c)n$ (where $c>0$ is a constant) leaf nodes, it evolves the exact target function in $O(\Tmax n \log^2 n)$ iterations in
expectation. While using the complete truth table to evaluate program quality requires
exponential time, we consider this setting for two main reasons: first, 
this setting represents the best-case model of the GP system's behaviour
(i.e. a system unable to find the optimal solution when given access to a reliable
fitness function is unlikely to be able to perform well with a noisy one);
and second,
the deterministic fitness values somewhat simplify the behaviour of the algorithm
and hence our analysis.

Afterwards we consider more realistic training sets of polynomial size sampled in each iteration uniformly at random from the complete truth table.
In practice some information about the function class to be evolved may be used to decide which inputs to use in the training set.
For instance, if the target function was known to be the conjunction of $n$ variables, then a compact training set of linear size would suffice to evolve the exact solution efficiently~\cite{LissovoiO18}.
However, we assume that the target function is an unknown arbitrary function composed of conjunctions and disjunctions of $n$ variables.
Our aim is to estimate the quality of the solution produced by the \RLSGP in this setting.

We show that with probability $1 - O(\log^2(n)/n)$ \RLSGP is
able to construct and return a conjunction with
a polynomially small generalisation error in a logarithmic number of iterations.
Hence, if multiple runs of the GP are performed as in practice, a solution that generalises well is
generated with probability converging quickly to 1 with the number of runs.

To achieve our results, we introduce a super-multiplicative drift theorem that makes use of a stronger drift than the linear one required by the traditional multiplicative drift theorem~\cite{DoerrJW12}. This new contribution to the portfolio of methodologies for the analysis of randomised search heuristics~\cite{OlivetoY11runtime,LehreO17bookchapter} allows for the achievement of drastically smaller bounds on the expected runtime in the presence of a strong multiplicative drift.

We complement our theoretical results with an empirical investigation that, on one hand, confirms
our theoretical intuition that leaf-only deletion may get stuck on local optima if a limit on the 
tree size is imposed for bloat control reasons. On the other hand, while the experiments indicate that the algorithm would evolve the solution more quickly without a limit on the tree size, the size limit reduces the amount of expected undesired binary disjunction operators in the final solution.

\section{Preliminaries} \label{sec:Prelim}

In this work, we will analyse the performance of the simple \RLSGP
algorithm on the \andn problem: evolving a conjunction of all $n$
input variables while using $F = \{\textsc{AND}, \textsc{OR}\}$ binary
functions and $L = \{x_1, \ldots, x_n\}$ input variables. When
program quality is evaluated using the complete truth table, the fitness
function $f(X)$ counts the number of truth-value assignments on which the
candidate solution $X$ differs from the target function $\hh(\mathbf{x}) = 
\andn = x_1 \wedge \ldots \wedge x_n$. From \cite{LissovoiO18}, we repeat
the observation that a conjunction of $a$ distinct variables differs from
\andn on $2^{n-a}-1$ rows of the complete truth table.

We will analyse the performance of the \RLSGP algorithm, which repeatedly
chooses the best between its current solution and an offspring generated by
applying the HVL-Prime mutation operator, which with equal probability
inserts, deletes, or substitutes a leaf node in the current solution
\cite{DurrettNO11}. We observe that the presence of disjunctions in the current solution 
may lead to bloat issues: each OR increases the minimum number of leaf nodes
required to represent the exact conjunction (up to a factor of at most 2,
depending on its position within the tree), can be difficult for
HVL-Prime to remove (as its deletion sub-operation only removes a single leaf
node and its immediate ancestor), and may additionally slow the progress toward
the optimum (as insertions under an OR have a diminished effect on the overall
solution semantics). To counteract this, we add a simple bloat control mechanism
to \RLSGP, making it reject trees which contain more than $\ell$ leaf nodes,
as described in Algorithm~\ref{alg:RLS-GP}.

\begin{algorithm}[t]
	\begin{algorithmic}[1]
		\State Initialise an empty tree $X$
		\For{$t \gets 1, 2, \ldots$}
			\State $X' \gets \text{HVL\text-Prime}(X)$
			\If{$\text{LeafCount}(X') \leq \Tmax$ \textbf{and} $f(X') \leq f(X)$}
				\State $X \gets X'$
			\EndIf
		\EndFor
	\end{algorithmic}
	\caption{The \RLSGP algorithm with a tree size limit $\Tmax$.} \label{alg:RLS-GP}
\end{algorithm}

With the tree size limit in place, applying the original HVL-Prime mutation operator
\cite{DurrettNO11} may cause \RLSGP with the limit $\ell$ to get stuck on
a local optimum.

\begin{theorem} \label{thm:rls-stuck}
The expected optimisation time of \RLSGP with leaf-only
deletion and substitution sub-operations of HVL-Prime,
and any $\Tmax > 0$ on \andn with $F=\{\text{AND}, \text{OR}\}$ is
infinite.
\end{theorem}
\begin{proofinappendix}
It is possible for \RLSGP to construct trees which
cannot be further improved by local mutations. One example of this is a tree
constructed by initially creating a disjunction of $\Tmax/2$ $x_1$ leaf nodes,
and then transforming each $x_1$ leaf into an $x_1 \wedge x_2$ subtree. No leaf
node in the final tree can be deleted or substituted without decreasing
fitness, and no insertion will be accepted due to the tree size limit,
rendering \RLSGP unable to reach the optimum. As this tree can be
constructed with non-zero probability, the expected time to construct the
optimal solution is infinite by the law of total expectation.
\end{proofinappendix}

To avoid this issue, we modify the deletion operation of HVL-Prime to allow
deletion of subtrees as described in Algorithm~\ref{alg:HVLPrime}.

\begin{algorithm}[t]
	\begin{algorithmic}[1]
		\State Choose $op \in \{\text{INS}, \text{DEL}, \text{SUB}\}$, $l \in L$, $f \in F$ uniformly at random
		\If{$X$ is an empty tree}
			\State Set $l$ to be the root of $X$.
		\ElsIf{$op = \text{INS}$}
			\State Choose a node $x \in X$ uniformly at random
			\State Replace $x$ with $f$, setting the children of $f$ to be $x$ and $l$, order chosen u.a.r.
		\ElsIf{$op = \text{DEL}$} \Comment{modified (subtree) deletion}
			\State Choose a node $x \in X$ uniformly at random
			\State Replace $x$'s parent in $X$ with $x$'s sibling in $X$
		\ElsIf{$op = \text{SUB}$}
			\State Choose a leaf node $x \in X$ uniformly at random
			\State Replace $x$ with $l$.
		\EndIf
		\State \Return the modified tree $X$
	\end{algorithmic}
	\caption{HVL-Prime with subtree deletion on tree $X$.} \label{alg:HVLPrime}
\end{algorithm}

We use the term \textit{sampled error} to refer to the fitness value of a particular solution
in a particular iteration, and \textit{generalisation error} to refer to the probability that
a particular solution is wrong on an input chosen uniformly at random from the set of all
$2^n$ possible inputs. When program quality is evaluated using the complete truth table,
the sampled error of a solution is always exactly $2^n$ times its generalisation error.
When the complete truth table is used, the goal of the GP system is to construct a solution that is
semantically equivalent to the target function i.e., achieve a sampled (and generalisation)
error of 0.

As it is computationally infeasible to evaluate all $2^n$ possible inputs for larger
values of $n$, we also analyse the behaviour of \RLSGP when evaluating solution quality
based on $s \in \mathrm{poly}(n)$ inputs chosen uniformly at random from the set of
all possible inputs. A fresh set of $s$ inputs is chosen in each iteration, and
$f(X)$, or the \textit{sampled error},
then refers to the number of inputs, among the chosen $s$, on which $X$
differs from the target function. The sampled error is thus a random variable,
and its expectation is exactly $s$ times the generalisation error of the solution. We bound
the probability of the sampled error deviating from its expectation in
Lemma~\ref{lem:sampled-error} below.
When a polynomial training set is used to evaluate program quality,
the goal of the GP system is to construct a solution with a low generalisation error.
On \andn, and most other non-trivial problems, we do not expect the GP systems to
reach a generalisation error of 0 while $s$ remains polynomial with respect to the
problem size, unless the problem's fitness landscape is well understood and a
problem-specific training set is used. We assume that this is not the 
case, and that the aim is to find a solution that has a polynomially small generalisation error.

\begin{lemma} \label{lem:sampled-error}
Let $s \in \mathrm{poly}(n)$ be the number of inputs sampled by the GP system,
$F$ be the generalisation error of a solution, and $X$ be a random variable
denoting the sampled error of that solution. Then, for any $c$ that is at
least a positive constant,
$$ |F s - X| \le \max\{c \lg n, F s \} $$
with probability at least $1 - n^{-\Omega(c)}$.
\end{lemma}

\begin{proofinappendix}
$X$ is a sum of $s$ Bernoulli variables, each
with a probability $F$ of assuming the value $1$ (and $0$ otherwise), and hence
$E[X] = Fs$. As both $X$ and $Fs$ are non-negative, $F s - X \leq F s$,
and we focus solely on the case where $X$ significantly exceeds its expectation,
the probability of which can be bounded by applying a Chernoff bound.

Suppose that $E[X] \geq (c/2) \lg n$; then, $\Pr[X \geq (1+1) E[X]] \leq e^{-E[X]/3} \leq n^{-\Omega(c)}$; and hence $|Fs - X| < Fs$, with probability at least $1-n^{-\Omega(c)}$. Otherwise, we upper bound $E[X] \leq \mu^+ = (c/2) \lg n$, and
apply a Chernoff bound using $\mu^+$ \cite[Theorem 66]{Doerr18tools}, obtaining $\Pr[X \geq (1+1) \mu^+] \leq e^{-\mu^+/3} = n^{-\Omega(c)}$; and hence $|Fs - X| \leq X \leq c \lg n$ with probability at least $1-n^{-\Omega(c)}$.
\end{proofinappendix}

Finally, we use the following notation throughout the paper: $\N := \{0,1,2,
\dots\}$, $\lg(n)$ and $\ln(e)$ denoting the base 2 and the natural logarithms 
of $n$, while $\log n$ is used in asymptotic bounds.

\section{Complete truth table} \label{sec:CTT}
In this section, we will present a runtime analysis of the \RLSGP algorithm
with subtree deletion (i.e., Algorithm~\ref{alg:RLS-GP})
on the \andn problem, using the complete truth table to evaluate solution
quality, i.e. executing each constructed program on all $2^n$ possible inputs.

\begin{theorem} \label{thm:lower-and}
The expected runtime of \RLSGP with $\Tmax \geq n$ on \andn is $E[T] = \Omega(n \log n)$.
\end{theorem}
\begin{proofinappendix}
No tree which does not contain all $n$ distinct variables can be
equivalent to the \andn function.
By a standard coupon collector argument, $\Omega(n \log n)$ insertion or
substitution operations are required to insert all $n$ distinct variables into
the tree.
\end{proofinappendix}

The following drift theorem deals with the situation that the expected progress when in distance $d$ from the target is of order $\Omega(d \log d)$. This assumption is slightly stronger than the linear, that is, $\Omega(d)$, progress assumed in the multiplicative drift theorem. Despite this apparently small difference, the resulting bounds for the expected time to reach the target differ drastically. For an initial distance of $d_0$, they are, roughly speaking, $O(\log d_0)$ for the multiplicative drift situation and $O(\log \log d_0)$ for our super-multiplicative drift.

\begin{theorem}[super-multiplicative drift theorem]\label{thm:smdrift}
  Let $\gamma > 1$ and $\delta > 0$.  Let $X_0, X_1, \dots$ be random variables taking values in $\Omega = \{0\} \cup [1,\infty)$. Assume that for all $t \in \N$ and all $x \in \Omega \setminus \{0\}$ such that $\Pr[X_t = x] > 0$ we have 
  \begin{equation}
  E[X_t - X_{t+1} \mid X_t = x] \ge (\log_\gamma(x)+1) \delta x. \label{eq:superdrift}
  \end{equation}
  Then the first hitting time $T = \min\{t \in \N \mid X_t = 0\}$ of zero satisfies \[E[T \mid X_0] \le \frac 3 \delta + \frac{2 (2+\log_2 \log_\gamma \max\{\gamma,X_0\}) \ln \gamma}{\delta}
.\]
\end{theorem}

\begin{proofinappendix}
  For all $k \in \N_{\ge 1}$, let $T_k := \min\{t \in \N \mid X_t < \gamma^{2^{k-1}}\}$. We first show that 
  \begin{equation*}
  E[T_{k} - T_{k+1}] \le \frac{1+2^k \ln \gamma}{(2^{k-1}+1) \delta}
  \end{equation*}
  holds for all $k \ge 1$. To this aim, we regard the process $Y_t$ defined for all $t \in \N$ by $Y_t = X_t$ if $t \le T_k-1$ and $Y_t = 0$ otherwise. By definition, $T^Y_k := \min\{t \in \N \mid Y_t < \gamma^{2^{k-1}}\}$ satisfies $T^Y_{k} = T_{k}$. The process $(Y_t)$ satisfies the multiplicative drift condition 
  \[E[Y_t - Y_{t+1} \mid Y_t] \ge (2^{k-1}+1) \delta Y_t.\]
  This follows from treating separately the trivial case $Y_t =0$ and the more interesting case $Y_t \ge \gamma^{2^{k-1}}$ and exploiting $Y_{t+1} \le X_{t+1}$, $Y_t = X_t$, and~\eqref{eq:superdrift} in the latter case.
  
  Let $T^Y := \min\{t \in \N \mid Y_t = 0\}$. Since $T^Y = T^Y_k = T_k$ and since $Y_t \le \gamma^{2^k}$ for all $t \ge T_{k+1}$, the multiplicative drift theorem~\cite{DoerrJW12} yields $E[T_{k} - T_{k+1}] = E[T^Y - T^Y_{k+1}] \le \frac{1+\ln \gamma^{2^k}}{(2^{k-1}+1)\delta} = \frac{1+2^k \ln \gamma}{(2^{k-1}+1)\delta}$. 
  
  By a simple application of the multiplicative drift theorem, we also observe that $E[T - T_1] \le \frac{1+\ln \gamma}{\delta}$.
  
  In the following, we condition on the initial value $X_0$. Assume that $X_0 \in [\gamma^{2^{k-1}},\gamma^{2^{k}})$ for some $k \in \N_{\ge 1}$. Then $T_{k+1} = 0$ and thus $T = \sum_{i=1}^k (T_i - T_{i+1}) + (T - T_1)$. We compute
  \begin{align*}
  E[T] & = \sum_{i=1}^k E[T_i - T_{i+1}] + E[T - T_1]
  \le \sum_{i=1}^k \frac{1+2^i \ln \gamma}{(2^{i-1}+1) \delta} + \frac{1+\ln\gamma}{\delta}\\
  &\le \frac 3 \delta + \frac{2(k+1)\ln \gamma}{\delta}
  \le \frac 3 \delta + \frac{2 (2+\log_2 \log_\gamma X_0) \ln \gamma}{\delta}.
  \end{align*}
  For $X_0 < \gamma$, we have in an analogous way $E[T] \le \frac{1 + \ln(X_0)}{\delta} \le \frac{1 + \ln \gamma}{\delta}$. This proves the claim.
\end{proofinappendix} 

The proof of the above theorem estimates the super-multiplicative drift by piece-wise multiplicative drifts. We preferred this proof method because of its simplicity and because it could, by using the multiplicative drift theorem with tail-bounds~\cite{DoerrG13algo}, also lead to tail-bounds for super-multiplicative drift as well (we do not elaborate on this as we do not need tail bounds). An alternative approach which would improve the time bound by a constant factor (again a feature we are not interested in here) would be to use variable drift~\cite{MitavskiyRC09,Johannsen10}.

We use the super-multiplicative drift theorem to prove our upper bound for the runtime of \RLSGP on the \andn function. We start by bounding the time spent in iterations in which the tree is not full, that is, it has not reached the size limit of having $\ell$ leaf nodes.

\begin{lemma} \label{lem:unfull-runtime}
Consider a run of \RLSGP on \andn, using a tree size limit of $\Tmax \geq n$.
Let $T$ be the number of iterations before the optimum is found, and $T_0 \leq
T$ be the number of these iterations in which the parent individual is not a
full tree. Then, $E[T_0]=O(\Tmax \, n \log^2 n)$.
\end{lemma}

\begin{proofinappendix}
To bound $E[T_0]$, we will apply Theorem~\ref{thm:smdrift} using solution
fitness as the potential function, and considering only the iterations in which
the tree is not full. While the tree \emph{is} full, we instead rely on the
elitism of the \RLSGP algorithm to not accept mutations which increase the
potential function value (i.e., offspring with a worse fitness value). Thus,
the $T_0$ iterations in which the tree is not full need not be contiguous.

In an iteration starting with a tree containing less than $\Tmax$ leaf nodes, it is possible to insert a new leaf node $x_i$ with an $AND$ parent anchored at the root of the tree.  We call such an operation a root-and. The probability that in one iteration a root-and with a fixed variable $x_i$ is performed, is at least $\frac 13 \cdot \frac 12 \cdot \frac 1 {2\ell} \cdot \frac 1n = \frac{1}{12\ell n}$.

We compute the expected fitness gain caused by such modifications. Because the fitness never worsens, it suffices to regard certain operations that improve the fitness. Recall further that the fitness is just the number of assignments to the variables $x_1, \dots, x_n$ such that the tree evaluates differently from \andn (``contradicting assignments'').

Let $x_1, \dots, x_n$ be such an assignment. This implies that not all $x_i$ are true, because any tree generated by \RLSGP evaluates correctly to true for the all-true assignment. Assume that exactly $k \ge 1$ of the variables $x_1, \dots, x_n$ are false, but that our tree solution evaluates to true. Then there are exactly $k$ variables such that a root-and with one of them would make this assignment evaluate to false (and thus improve the fitness since this assignment is not contradicting anymore). The probability for such a mutation is at least $\frac{k}{12 \ell n}$.

For any $1 \leq i \leq n$, there are exactly $\binom{n}{i}$ assignments where
exactly $i$ variables are set to false, and hence there are exactly
$\sum_{i=1}^{k-1} \binom{n}{i}$ possible assignments where less than $k$
variables are set to false. Therefore, if the fitness of the current solution is at
least $M_k = 2 \sum_{i=1}^{k-1} \binom{n}{i}$, at least half of the assignments
contributing to the fitness have at least $k$ variables set to false.
Only regarding the progress caused by these, we have, for $x \ge M_k$,
\begin{gather}
E\left[f(X^t) -f(X^{t+1}) \mid f(X^{t}) = x\right] \geq \frac{1}{12\Tmax} \, \frac{k}{n} \, x. \label{eq:k-drift}
\end{gather}

Since for $n$ sufficiently large we have $M_k \le 2 n^{k-1}$ for all $k \in [1..n]$. This implies that for all $x \in [1..2^n]$ and all $t \in \N$, we have
\begin{align*}
E\left[f(X^t) -f(X^{t+1}) \mid f(X^{t}) = x\right] 
&\ge \frac{1}{12 \ell n} (\lfloor \log_n(x/2) \rfloor+ 1) x \\
&\ge \frac{1}{36 \ell n} (\log_n(x)+1) x,
\end{align*}
where the last estimate uses $n \ge 2$. Hence Theorem~\ref{thm:smdrift} with $\gamma=n$ and $\delta = 1/36\ell n$ gives \[E[T] \le 36 \ell n (3+ 2(2 + \log_2 \log_n 2^n))\ln n) = O(\ell n \log^2 n).\]
\end{proofinappendix}

We can then show that the conditions required to apply Lemma~\ref{lem:unfull-runtime}
occur sufficiently often to not affect the asymptotic expected runtime.

\begin{theorem} \label{thm:CTT-runtime}
Consider a run of \RLSGP on \andn, using a tree size limit of $\Tmax = (1+c) n$.
Let $T$ be the number of iterations before the optimum is found. If
$c = \Theta(1)$, then $E[T] = O(\Tmax \, n \log^2 n)$.
\end{theorem}
\begin{proofinappendix}
To prove the theorem, we combine the result of Lemma~\ref{lem:unfull-runtime}
with an argument showing that with high probability, the parent solution
contains fewer than $\ell$ leaf nodes in at least a constant fraction of any $t
\in \Omega(\Tmax\, n \log^2n)$ iterations.

Let $T' = c^* \Tmax n \, \log^2 n$, for some constant $c^* > 0$, be an upper
bound on the expected number of iterations $E[T_0]$ in which the tree is not 
full
before the optimum solution is found per Lemma~\ref{lem:unfull-runtime}.
By an application of Markov's inequality, the
probability that the optimum is found in at most $2T'$ such iterations is at
least $1/2$. We will show that if $\Tmax = (1+c) n$, for any constant $c > 0$,
$2T'$ such iterations occur in $(2+c')T'$ iterations with high
probability, where $c' > 0$ is constant with respect to $n$. The theorem
statement then follows from a simple waiting time argument: during each period
of $(2+c')T'$ iterations, the optimum is found with at least probability $1/2
\cdot (1 - o(1)) = \Omega(1)$, so the expected number of such periods before the
optimum is found is at most $O(1)$, and thus the expected runtime is at most
$O(T') = O(\Tmax n \, \log^2 n)$ iterations.

We will now show that during any $N \in \Omega(\Tmax\, n \log^2n)$ iterations,
with high probability and for some constant $c'' > 0$, deletions of at least
$c'' N$ leaf nodes in total will be accepted. 
As each iteration can at most increase the number of leaf nodes in the tree
by 1, there will with high probability be
at least $c'' N$ iterations in which the tree is not full among any $(1+c'') N$
iterations. As $T' \in \Omega(\Tmax\, n \log^2n)$, $2T'$ iterations in which
the tree is not full will with high probability occur in $(2+c')T'$ iterations
where $c' = 2/c'' = \Omega(1)$.

Consider a tree $X$ with exactly $\Tmax$ leaf nodes. Let $L_A(X)$ be a set of
leaf nodes connected to the root of $X$ via only AND nodes, and call
\emph{essential} all the leaf nodes in this set that contain a variable which
only appears on nodes in this set exactly once. If $X$ is non-optimal, at most
$n-1$ leaf nodes in $X$ are essential, and at least $\Tmax - (n-1)$ leaf nodes
are non-essential. All
non-essential nodes are either directly deletable (in the case of redundant
copies of variables in $L_A(X)$), or indirectly deletable (by deleting a branch
at any of their OR ancestors).

Every non-essential leaf node can thus be deleted by performing an HVL-Prime
deletion sub-operation on at least one node in the tree.
For some non-essential leaf nodes, a larger subtree may need to be deleted to
remove the leaf without adversely impacting fitness. The longer waiting time
for such subtree deletions (requiring that the root of the subtree be chosen
for deletion rather than one of the many leaf nodes in the subtree) is balanced
by the increased number of leaf nodes deleted as part of the mutation.
We note that the tree contains $2\Tmax-1$ nodes, and thus for $\Tmax \geq
(1+c)n$ and any $c > 0$, an HVL-Prime mutation in expectation reduces the
number of leaf nodes in the tree by at least 
$$ E[\Delta] \geq \frac{1}{3}\frac{\Tmax - (n - 1)}{2\Tmax - 1} \geq
\frac{\Tmax - n}{6\Tmax} \geq \frac{c}{6+6c} \geq \delta \in \Omega(1),$$
where $\delta > 0$ is a positive constant, as $c \in \Omega(1)$.

Let $X_1, \ldots, X_N$ be the number of leaf nodes deleted in an accepted
mutation during each iteration performed while the tree is full, and $X =
\sum_{i=1}^{N} X_i$. Furthermore, define a sequence $Z_0, \ldots, Z_N$, where
$Z_0 := 0$ and $Z_i := Z_{i-1} + X_i - \delta$; clearly, $Z_N - Z_0 = Z_N = X -
\delta N$. We will show that $Z_N > - \delta N/2$ (and therefore $X >
\delta N/2 \in \Omega(N)$) holds with high probability.

As $E[Z_{i} \mid Z_1, \ldots Z_{i-1}] = Z_{i-1} + E[X_{i} \mid Z_1, \ldots
Z_{i-1}] - \delta \geq Z_{i-1}$, the sequence $Z_0, \ldots, Z_N$ is a
sub-martingale, and $c_i := |Z_i - Z_{i-1}| \leq \Tmax$. Hence, by applying the
Azuma-Hoeffding inequality for $N \in \Omega(\Tmax\, n \log^2n)$ and $t =
\delta N/2$,
\begin{align*}
\Pr[Z_N - Z_0\leq - t] & \leq \exp\left(\frac{-t^2}{2\sum_{i=1}^{N} c_i^2}\right)
\leq \exp\left(\frac{-\delta^2 N}{8 \Tmax^2}\right)
\\ & \leq n^{-\Omega(\log n)}
\end{align*}
as $N/\Tmax^2 = \Omega(n \Tmax \log^2 n / \Tmax^2) = \Omega(\log^2 n)$ for
$\Tmax = (1+c)n$ where $c$ is a constant.

Thus, there exists a constant $c'' > 0$ such that over the course of $N \in
\Omega(n \Tmax \log^2 n)$ iterations where the tree is full, deletions of at
least $\delta N/2 = c'' N$ leaf nodes are accepted with high
probability, and hence over the course of $2/c'' N$ iterations, at least
$2N$ iterations occur while the tree is not full with high probability.
Setting $N = T' = c^* n \Tmax \log^2 n$ iterations per
Lemma~\ref{lem:unfull-runtime} completes the proof: among $\Theta(T')$
iterations, at least $\Omega(T')$ will take place while the tree is not full,
allowing the application of the Markov inequality and waiting time arguments to
produce the bound on the expected runtime.
\end{proofinappendix}

\section{Polynomially sized training sets} \label{sec:Incomplete}

While Theorem~\ref{thm:CTT-runtime} provides a polynomial bound
on the number of iterations required to evolve the conjunction of $n$ variables,
calculating solution quality by evaluating the output of the candidate solution
and the target function on each one of the $2^n$ possible inputs in each iteration
requires exponential computational effort, and is thus only computationally feasible
for relatively modest values of $n$.

In this section, we consider the behaviour of the \RLSGP algorithm when using only
a polynomial computational effort in each iteration. To this end, the solution
quality is compared by evaluating the output of the ancestor solution, the offspring,
and the target function on only a polynomial number of inputs (``the training set''),
sampled uniformly at random from the set of all possible inputs in each iteration.
This setting was previously considered in \cite{LissovoiO18}, where it was shown that the
\RLSGP and the (1+1)~GP algorithms using $F = \{AND\}$ are able to construct a solution
with $O(\log n)$ distinct variables which fits a polynomially large training set in polynomial time.

For our main theoretical result below, we opt to have \RLSGP terminate and return
a solution once the sampled error on the training set is below a logarithmic acceptance
threshold. This effectively prevents \RLSGP from entering a region of the search
space where the mechanism it uses to evaluate program quality is overly noisy. This
slightly decreases the expected solution quality, but does preserve the
overall guarantee on the quality of the produced solution.

\begin{theorem} \label{thm:incomplete-single-run}
For any constant $c > 0$, an instance of the
\RLSGP algorithm with $F = \{AND, OR\}$, $L = \{x_1, \ldots, x_n\}$, $\Tmax \geq n$,
using a training set of $s = n^c \lg^2 n$ rows sampled uniformly
at random from the complete truth table in each iteration to evaluate solution
quality, and terminating when the sampled error of the solution is at most
$c' \lg n$, where $c'$ is an appropriately large constant,
will with probability at least $1 - O(\log^2(n)/n)$ terminate within
$O(\log n)$ iterations, producing a solution with a generalisation
error of at most $n^{-c}$.
\end{theorem}
To prove this theorem, we will show that \RLSGP is able to create a tree that
contains no more than one copy of each variable, no OR functions, and enough
distinct variables to sample an error below the acceptance threshold within
$O(\log n)$ iterations with probability at least $1 - O(\log^2(n)/n)$.
Additionally, we will show that with high probability, the GP system will not
terminate early (i.e., it will not return a solution with a generalisation error greater 
than $n^{-c}$).

\begin{lemma} \label{lem:incomplete-happy-runtime}
If \RLSGP never accepts solutions containing OR nodes or multiple copies of
any variable, and never accepts solutions with a worse generalisation error than
their ancestors, it will within $O(\log n)$ iterations reach a solution with a
sampled error below $c' \lg n$, where $c' > 0$ is an appropriate constant,
with probability at least $1 - O(1/n)$.
\end{lemma}
\begin{proofinappendix}
To ensure that an error below $c' \lg n$ is sampled, we consider the time required
to construct a solution with an expected sampling error of at most $(c'/4) \lg n$.
Such a sampling error can be achieved by a generalisation error of
at most $((c'/4) \lg n) / (n^c \lg^2 n) = (c'/4) n^{-c} / \lg n \geq n^{-(c+1)}$
(for a sufficiently large $n$), i.e., a conjunction of $(c+1)\lg n$ variables or more.

The time required to construct such a conjunction under the lemma's conditions
can be bounded by lower-bounding the probability of inserting a new variable connected
to the tree using an AND node, and using a Chernoff bound to show that a sufficient
number of such insertions occur within a particular number of iterations (as the
number of distinct variables in the current solution is never reduced by the lemma's
conditions). Specifically, suppose that the current solution contains $i < n/2$ 
distinct variables and no OR nodes, and let $X_i$ be the event that a mutation inserts
a new variable and connects it to the tree using an AND node, and is accepted. We bound
$ \Pr[X_i] \geq (1/3) (1/2) (n-i)/n \geq \delta $, i.e.,
$\delta \geq 1/12$ for $i < n/2$. The probability that
at least $(c+1)\lg n$ such mutations are accepted within $(c''/\delta) (c+1) \lg n = O(\log n)$
iterations is then, by applying a Chernoff bound \cite[Lemma 1.18]{Doerr11bookchapter}, at
least $1 - e^{-\Omega(c'' \log n)} = 1 - n^{-\Omega(c'')}$.
Thus, when $c''$ is a sufficiently large constant, this probability is at least
$1 - O(1/n)$.

We bound the probability that a solution with a low-enough expected sampled error
does not meet the acceptance threshold by applying Lemma~\ref{lem:sampled-error}:
once a solution with an expected sampled error of at most $(c'/4) \lg n$ is 
constructed, the probability that its sampled error exceeds the acceptance threshold
is at most $n^{-\Omega(c')}$, and thus, when $c'$ is picked appropriately, the solution
is accepted immediately with probability at least $1-O(1/n)$.

By combining the failure probabilities using a union bound, we conclude that
\RLSGP under the conditions of the lemma and with an appropriately-chosen
constant $c'$, is able to construct a
solution with an acceptable sampled error within $O(\log n)$ iterations with
probability at least $1-O(1/n)$.
\end{proofinappendix}

We will now use this bound on the runtime of \RLSGP to show that it is likely
to avoid all of the potential pitfalls preventing the application of
Lemma~\ref{lem:incomplete-happy-runtime}.

\begin{lemma} \label{lem:incomplete-no-bad-things}
With probability at least $1-O(\log^2(n)/n)$, during its first $O(\log n)$
iterations and while the expected sampled error of its current solution remains
above $(c'/4)\lg n$, \RLSGP is able to avoid accepting mutations which: (1)
insert copies of a variable already present in the current solution, (2) insert
OR nodes, or (3) increase the generalisation error of the current solution.
\end{lemma}
\begin{proofinappendix}
For claim (1), we note that within the first $O(\log n)$ iterations, the tree
will contain at most $O(\log n)$ distinct variables (as each iteration of 
\RLSGP is only able to insert one additional variable). Thus, the probability
that a mutation operation adds a variable which is already present in the
solution (using either the insertion or substitution sub-operation of
HVL-Prime) is at most $O(\log n / n)$,
and by a union bound, this does not occur during the first $O(\log n)$
iterations with probability at least $1 - O(\log^2 (n) / n)$.

For claim (2), we note that there are two main ways an OR can be introduced into
the solution by an insertion operation: either the OR is semantically neutral
(which, if the ancestor contains only ANDs and unique variables requires replacing
a leaf $x_i$ with $x_i \vee x_i$), or the sampling process used to evaluate solution
fitness did not sample any inputs on which the offspring is wrong and the ancestor
is correct. We will consider the two possibilities separately.

As semantically-neutral insertions of OR nodes require inserting a
duplicate copy of a variable, claim (1) already provides the desired probability
bound on these insertions not occurring within $O(\log n)$ iterations
(and hence not being accepted).
All other OR insertions will increase the generalisation error of the solution. The
magnitude of this increase depends on the number of distinct
variables in the subtree displaced by the insertion, with insertions displacing
only a single leaf node being the easiest to accept.

If a leaf of the ancestor solution is replaced with a disjunction with a new variable,
we use the term \textit{witness} to refer to inputs which set the displaced variable
to $0$ while setting the remaining variables in the offspring solution to $1$. 
As the offspring solution also differs from the target function on all the inputs on which
the ancestor solution does so, as long as
the sampling procedure samples at least one witness, \RLSGP will reject the mutated
solution. Suppose the ancestor conjunction contains $U$ distinct variables;
it is then incorrect on $2^{n-U}-1$ possible inputs, while there are at least
$2^{n-(U+1)}$ witnesses; i.e. the probability of randomly selecting a witness
is at least half that of randomly selecting a row on which the ancestor is wrong.
Thus, if the expected sampled error of the ancestor solution is at least $X$, 
the expected number of witnesses in the sample is at least $X/2$. By a Chernoff bound,
the probability that fewer than $(c'/16) \lg n$ witnesses are present in the sample is at most
$e^{-(c'/128) \lg n} = n^{-\Omega(c')}$. By setting the constant $c'$
appropriately, this probability can be made into $O(1/n)$, and by a union bound,
the probability that no OR which increases the generalisation error is accepted
within $O(\log n)$ iterations while the expected sampled error of the solution remains
above $(c'/4) \lg n$ is at least $1-O(\log(n)/n)$.

Finally, for claim (3), we note that decreasing the number of distinct variables
in the solution more than doubles its generalisation error. Applying a similar
argument for rejecting detrimental ORs above (this time, the expected
number of witnesses in the sample is at least $X$),
the probability that no mutations increasing the generalisation error are accepted
during $O(\log n)$ iterations is at least $1-O(\log(n)/n)$.

Combining the error probabilities of the three claims using a union bound yields
the theorem statement.
\end{proofinappendix}

Finally, we show that with high probability, \RLSGP does not
terminate unacceptably early (i.e. by sampling an error below the acceptance
threshold for a solution with a worse generalisation error than desired by
Theorem~\ref{thm:incomplete-single-run}).

\begin{lemma} \label{lem:no-bad-solutions}
With high probability, no solution with a generalisation error greater than
$n^{-c}$ has a sampled error of at most $c' \lg n$ on a set of $s \geq
n^{c}\lg^2 n$ rows sampled random from the complete truth table, within any
polynomial number of iterations.
\end{lemma}
\begin{proofinappendix}	
Recall that when sampling $s$ rows uniformly at random from the complete truth
table to evaluate solution fitness, \RLSGP terminates and returns
the current solution when the solution appears wrong on at most $c' \lg n$ of
the sampled rows. As the generalisation error of a solution is also
the probability that the solution is wrong on a uniformly-sampled row of the
complete truth table, a solution $X$ with a generalisation error $g(X)$
of at least $n^{-c}$, has an expected sampled error $E(f(X)) \geq \lg^2 n$
on $s = n^{c}\lg^2 n$ rows sampled uniformly at random.
Applying a Chernoff bound, the probability that the
sampled error $Y$ is less than half of its expected value (which for
large-enough $n$ is above the $c' \lg n$ threshold), is super-polynomially
small: 
$$ \Pr\left[Y \leq 1/2 \, E[Y]\right] \leq e^{-E[Y]/8} \leq n^{-\Omega(\log n)}. $$

By a union bound, \RLSGP with high probability does not
return a solution with a generalisation error of at least $n^{-c}$ within any
polynomial number of iterations when sampling $s = \Omega(n^{c} \lg^2 n)$ rows
of the complete truth table uniformly at random to evaluate solution quality in
each iteration.
\end{proofinappendix}

Our main result is proved by combining these lemmas.

\begin{proofinappendix}[\proofOf Theorem~\ref{thm:incomplete-single-run}]
By Lemma~\ref{lem:incomplete-no-bad-things}, the conditions necessary to apply
Lemma~\ref{lem:incomplete-happy-runtime} occur with probability at least
$1-O(\log^2(n)/n)$, and thus with probability at least $1-O(\log^2(n)/n)-O(1/n)$,
a solution with a sampled error meeting the acceptance threshold will be
found and returned within $O(\log n)$ iterations.
By Lemma~\ref{lem:no-bad-solutions}, the generalisation error of any
solution returned by \RLSGP within a polynomial number of iterations is with
high probability better than the desired $n^{-c}$.
\end{proofinappendix}

We remark that performing $\lambda$ runs of \RLSGP, as is often done in practice,
and terminating once any instance determines that its current solution meets the
acceptance threshold, will guarantee that a solution with the desired generalisation
error is produced using $O(\lambda \log n)$ fitness evaluations with probability 
$1 - n^{-\Omega(\lambda)}$.

\section{Experiments} \label{sec:Experiments}

We performed experiments to complement our theoretical results.
For each choice of algorithm and problem parameters, we performed
500 independent runs of the GP system.

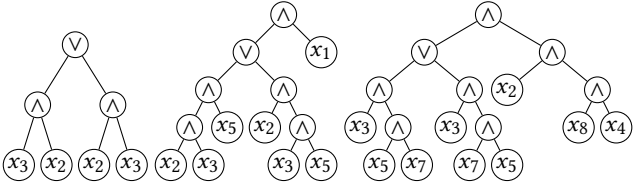
\begin{figure}[t]
	\hfill
	\begin{tikzpicture}[
		level 1/.style={sibling distance=10mm},
		level 2/.style={sibling distance=5mm}, 
		level 3/.style={sibling distance=5mm}, 
		level distance=8mm,
		every node/.style={draw,circle,inner sep=0.25mm}
	]
	\node {$\vee$}
child { node {$\wedge$}
child { node {$x_3$} }
child { node {$x_2$} } }
child { node {$\wedge$}
child { node {$x_2$} }
child { node {$x_3$} } }
	;
	\end{tikzpicture} \hfill
	\begin{tikzpicture}[
		level 1/.style={sibling distance=10mm},
		level 2/.style={sibling distance=10mm}, 
		level 3/.style={sibling distance=5mm}, 
		level distance=5mm,
		every node/.style={draw,circle,inner sep=0.25mm}
	]
	\node {$\wedge$}
child { node {$\vee$}
child { node {$\wedge$}
child { node {$\wedge$}
child { node {$x_2$} }
child { node {$x_3$} } }
child { node {$x_5$} } }
child { node {$\wedge$}
child { node {$x_2$} }
child { node {$\wedge$}
child { node {$x_3$} }
child { node {$x_5$} } } } }
child { node {$x_1$} };
	\end{tikzpicture} \hfill
    \begin{tikzpicture}[
		level 1/.style={sibling distance=17mm},
		level 2/.style={sibling distance=12mm}, 
		level 3/.style={sibling distance=5mm}, 
		level distance=5mm,
		every node/.style={draw,circle,inner sep=0.25mm}
	]
\node {$\wedge$}
child { node {$\vee$}
child { node {$\wedge$}
child { node {$x_{3}$} }
child { node {$\wedge$}
child { node {$x_{5}$} }
child { node {$x_{7}$} } } }
child { node {$\wedge$}
child { node {$x_{3}$} }
child { node {$\wedge$}
child { node {$x_{7}$} }
child { node {$x_{5}$} } } } }
child { node {$\wedge$}
child { node {$x_{2}$} }
child { node {$\wedge$}
child { node {$x_{8}$} }
child { node {$x_{4}$} } } };
\end{tikzpicture} \hspace*{\fill}
	
	\caption{Examples of locally optimal trees, which cannot be improved by substitution or have any single leaf deleted without affecting fitness, constructed by RLS-GP using leaf-only substitution and deletion operations.}
	\label{fig:leaf-bad-trees}
\end{figure}

Theorem \ref{thm:rls-stuck}
showed that using the standard HVL-Prime operator, which applies leaf-only deletion and substitution, can cause \RLSGP with the complete truth table to get stuck on a local
optimum when a tree size limit is imposed, thus leading to infinite expected runtime. However, the theorem does not provide bounds on the probability that this event occurs. 
Table~\ref{tab:leaf-RLS-CTT} summarises the experimental behaviour of \RLSGP.
The experiments confirm that when using small tree size limits, \RLSGP indeed gets stuck on local optima. Examples of the ones constructed during the runs are depicted in Fig. \ref{fig:leaf-bad-trees}. However, the probability of getting stuck decreases as $\Tmax$, the limit on the size
of the tree, increases.
Concerning solution quality, with small tree size limits, the number of redundant variables in the final solution decreases at the expense of higher runtimes. For $\Tmax=n$, {\it `exact'} solutions are returned when the algorithm does not get stuck. On the other hand, larger tree size limits (including no limit) lead to smaller expected runtimes at the expense of redundant variables in the final solutions.  

\begin{table}[t]
	\centering
	\setlength\tabcolsep{1mm}
	\begin{tabular}{|c|ccc|ccc|} \hline
    	& \multicolumn{3}{c}{$\Tmax = n$} & \multicolumn{3}{|c|}{$\Tmax = n+1$} \\
		\textbf{n} & $B$ & $\overline{T}$ & $\overline{S}$ & $B$ & $\overline{T}$ & $\overline{S}$ \\ \hline
 4 &	 0.008 &	46.3 (28.0) &	4.0 (0.0) &	0.002 &	40.9 (21.8) &	4.4 (0.5)	\\
 8 &	 0.002 &	151.8 (91.9) &	8.0 (0.0) &	0.004 &	113.8 (51.5) &	8.6 (0.5)	\\
 12 &	 0.016 &	284.1 (148.2) &	12.0 (0.0) &	0.002 &	214.3 (99.5) &	12.7 (0.5)	\\
 16 &	 0.008 &	469.9 (258.0) &	16.0 (0.0) &	0.010 &	345.8 (161.0) &	16.8 (0.4)	\\
 \hline
		\multicolumn{7}{c}{} \\ \hline
    	&  \multicolumn{3}{c}{$\Tmax = 2n$} & \multicolumn{3}{|c|}{$\Tmax = \infty$} \\
		\textbf{n} & $B$ & $\overline{T}$ & $\overline{S}$ & $B$ & $\overline{T}$ & $\overline{S}$ \\ \hline
 4 &	 0 &	42.5 (25.8) &	5.1 (1.2) &	0 &	38.9 (24.3) &	5.4 (2.0)	\\
 8 &	 0 &	98.8 (49.0) &	11.0 (2.3) &	0 &	95.3 (43.8) &	11.2 (3.0)	\\
 12 &	 0 &	170.7 (99.7) &	17.1 (3.3) &	0 &	160.1 (57.1) &	17.9 (4.5)	\\
 16 &	 0 &	232.5 (80.9) &	23.8 (4.1) &	0 &	235.3 (92.7) &	24.6 (6.0)	\\ \hline
	\end{tabular}
    
\caption{Proportion of runs stuck in a local optimum ($B$), and average runtime
($\overline{T}$) and solution size ($\overline{S}$) of successful runs of the
\RLSGP using leaf-only substitution and deletion with the complete truth table 
to evaluate solution quality for varying $n$ and $\Tmax$. Standard deviations
appear in parentheses.}
	\label{tab:leaf-RLS-CTT}
\end{table}

We now turn our attention to the 
HVL-Prime modified to allow subtree deletion, as considered by
Theorem~\ref{thm:CTT-runtime}. 
As predicted by the theory, \RLSGP never gets stuck in our
experiments when using the complete truth table and a tree size limit. Table~\ref{tab:tree-RLS-CTT} shows the average number of
iterations required to find the global optimum for various problem sizes
and varying tree size limits.
Once again the experiments show that smaller tree size limits lead to lower numbers of redundant variables at the expense of a higher runtime. Larger limits, including no limit at all, lead to faster runtimes at the expense of admitting more redundant variables. Noting that in practical applications a tree size limit is often necessary,
we leave the proof that the algorithm evolves an exact conjunction without any limits on the tree size for future work.

\begin{table}[t]
	\centering
	\setlength\tabcolsep{1.5mm}
	\begin{tabular}{|c|cc|cc|cc|cc|} \hline
    	& \multicolumn{2}{c|}{$\Tmax = n$} & \multicolumn{2}{c}{$\Tmax = n+1$} & \multicolumn{2}{|c}{$\Tmax = 2n$} & \multicolumn{2}{|c|}{$\Tmax = \infty$} \\
		\textbf{n} & $\overline{T}$ & $\overline{S}$ & $\overline{T}$ & $\overline{S}$ & $\overline{T}$ & $\overline{S}$ & $\overline{T}$ & $\overline{S}$ \\ \hline
 4 &	 51.2 & 4.0 &	 42.5 & 4.4 &	 38.8 & 5.1 &	 39.1 & 5.3 \\
 &	 (31.1) & (0.0) &	 (23.5) & (0.5) &	 (20.8) & (1.2) &	 (22.3) & (1.8) \\ \hline
 8 &	 147.5 & 8.0 &	 129.9 & 8.7 &	 93.5 & 11.3 &	 92.3 & 11.6 \\
 &	 (83.3) & (0.0) &	 (69.1) & (0.5) &	 (39.1) & (2.4) &	 (38.1) & (3.0) \\ \hline
 12 &	 325.9 & 12.0 &	 233.4 & 12.8 &	 153.6 & 17.7 &	 151.2 & 18.3 \\
 &	 (184.4) & (0.0) &	 (123.9) & (0.4) &	 (56.6) & (3.1) &	 (50.3) & (3.8) \\ \hline
 16 &	 544.6 & 16.0 &	 377.0 & 16.9 &	 228.3 & 24.5 &	 221.0 & 25.2 \\
 &	 (333.8) & (0.0) &	 (176.0) & (0.4) &	 (74.6) & (3.7) &	 (72.0) & (4.9) \\ \hline
	\end{tabular}
	\caption{
Average runtime ($\overline{T}$) and solution size ($\overline{S}$) of 
\RLSGP using the subtree deletion sub-operation, and the complete truth table 
to evaluate solution fitness, for varying $n$ and $\Tmax$. Standard deviations
appear in parentheses.
}
	\label{tab:tree-RLS-CTT}
\end{table}

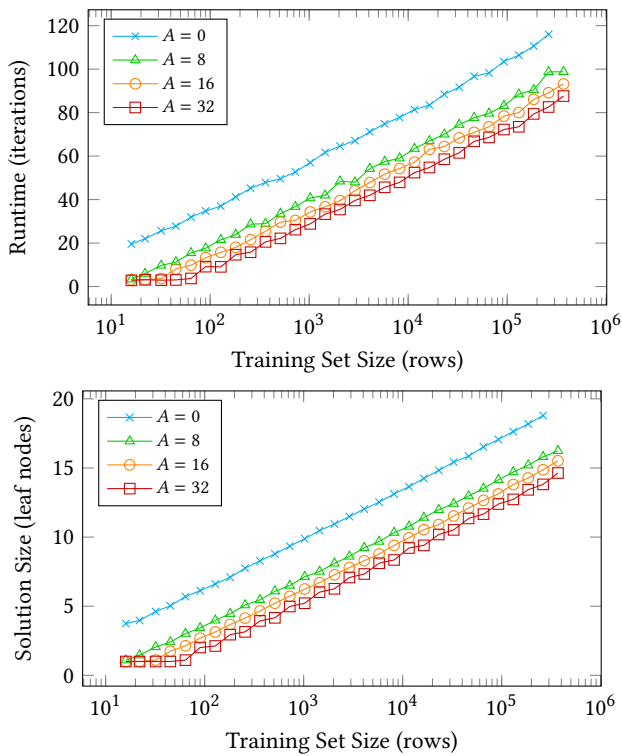
\begin{figure}[t]
	\pgfplotstableread{data/runtime-n50-linf-s0.tsv}\runFiftyInfZero
	\pgfplotstableread{data/treesize-n50-linf-s0.tsv}\sizeFiftyInfZero
    \pgfplotstableread{data/runtime-n50-linf-s16.tsv}\runFiftyInfSixteen
	\pgfplotstableread{data/treesize-n50-linf-s16.tsv}\sizeFiftyInfSixteen
    \pgfplotstableread{data/runtime-n50-linf-s8.tsv}\runFiftyInfEight
    \pgfplotstableread{data/treesize-n50-linf-s8.tsv}\sizeFiftyInfEight
    \pgfplotstableread{data/runtime-n50-linf-s32.tsv}\runFiftyInfThirtytwo
    \pgfplotstableread{data/treesize-n50-linf-s32.tsv}\sizeFiftyInfThirtytwo
	\begin{tikzpicture}
	  \begin{semilogxaxis}[
		  width=\linewidth, height=0.65\linewidth,
	    legend pos = north west,
		 xlabel={Training Set Size (rows)},
		 ylabel={Runtime (iterations)},
         try min ticks=5,
         legend cell align={left},
         legend style={nodes={scale=0.8, transform shape}}, 
	    ]
	    \addplot[cyan,mark=x] table[x index = {0}, y index = {2}]{\runFiftyInfZero};
        \addplot[green!80!black,mark=triangle] table[x index={0},y index={1}]{\runFiftyInfEight};
	    \addplot[orange,mark=o] table[x index = {0}, y index = {2}]{\runFiftyInfSixteen};
        \addplot[red!80!black,mark=square] table[x index={0},y index={1}]{\runFiftyInfThirtytwo};
	    \legend{$A=0$,$A=8$,$A=16$,$A=32$}
	  \end{semilogxaxis}
	\end{tikzpicture}\hfill
	\begin{tikzpicture}
	  \begin{semilogxaxis}[
		  width=\linewidth, height=0.65\linewidth,
	    legend pos = north west,
		 xlabel={Training Set Size (rows)},
		 ylabel={Solution Size (leaf nodes)},
         try min ticks=5,
         legend cell align={left},
         legend style={nodes={scale=0.8, transform shape}}, 
	    ]
	    \addplot[cyan,mark=x] table[x index = {0}, y index = {2}]{\sizeFiftyInfZero};
        \addplot[green!80!black,mark=triangle] table[x index={0},y index={1}]{\sizeFiftyInfEight};
	    \addplot[orange,mark=o] table[x index = {0}, y index = {2}]{\sizeFiftyInfSixteen};
        \addplot[red!80!black,mark=square] table[x index={0},y index={1}]{\sizeFiftyInfThirtytwo};
	    \legend{$A=0$,$A=8$,$A=16$,$A=32$}
	  \end{semilogxaxis}
	\end{tikzpicture}
	\caption{Average runtime and tree size produced by \RLSGP with subtree deletion, using an incomplete training set, stopping once sampled error is at most $A$, $n=50, \Tmax = \infty$, averaged over 500 independent simulations.}
	\label{fig:incomplete-training-results}
\end{figure}

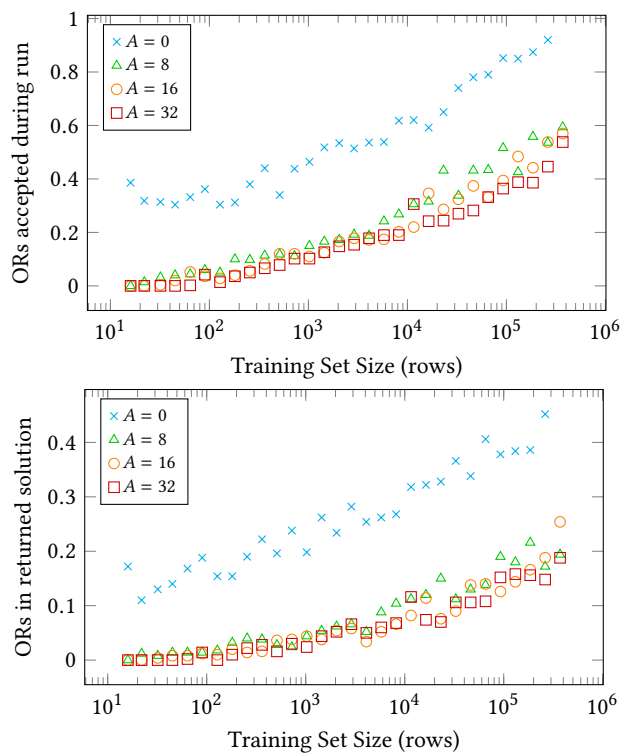
\begin{figure}[t]
	\pgfplotstableread{data/insOR-n50-linf-s0.tsv}\insFiftyInfZero
	\pgfplotstableread{data/finOR-n50-linf-s0.tsv}\finFiftyInfZero
    \pgfplotstableread{data/insOR-n50-linf-s16.tsv}\insFiftyInfSixteen
	\pgfplotstableread{data/finOR-n50-linf-s16.tsv}\finFiftyInfSixteen
    \pgfplotstableread{data/insOR-n50-linf-s8.tsv}\insFiftyInfEight
	\pgfplotstableread{data/finOR-n50-linf-s8.tsv}\finFiftyInfEight
	\pgfplotstableread{data/insOR-n50-linf-s32.tsv}\insFiftyInfThirtytwo
	\pgfplotstableread{data/finOR-n50-linf-s32.tsv}\finFiftyInfThirtytwo
	\begin{tikzpicture}
	  \begin{semilogxaxis}[
		  width=\linewidth, height=0.65\linewidth,
	    legend pos = north west,
		 xlabel={Training Set Size (rows)},
		 ylabel={ORs accepted during run},
         try min ticks=5,
         legend cell align={left},
         legend style={nodes={scale=0.8, transform shape}}, 
	    ]
	    \addplot[cyan,mark=x,only marks] table[x index = {0}, y index = {2}]{\insFiftyInfZero};
        \addplot[green!80!black,mark=triangle,only marks] table[x index={0},y index={1}]{\insFiftyInfEight};
	    \addplot[orange,mark=o,only marks] table[x index = {0}, y index = {2}]{\insFiftyInfSixteen};
        \addplot[red!80!black,mark=square,only marks] table[x index={0},y index={1}]{\insFiftyInfThirtytwo};
	    \legend{$A=0$,$A=8$,$A=16$,$A=32$}
	  \end{semilogxaxis}
	\end{tikzpicture}\hfill
	\begin{tikzpicture}
	  \begin{semilogxaxis}[
		  width=\linewidth, height=0.65\linewidth,
	    legend pos = north west,
		 xlabel={Training Set Size (rows)},
		 ylabel={ORs in returned solution},
         try min ticks=5,
         legend cell align={left},
         legend style={nodes={scale=0.8, transform shape}}, 
	    ]
	    \addplot[cyan,mark=x,only marks] table[x index = {0}, y index = {2}]{\finFiftyInfZero};
        \addplot[green!80!black,mark=triangle,only marks] table[x index={0},y index={1}]{\finFiftyInfEight};
	    \addplot[orange,mark=o,only marks] table[x index = {0}, y index = {2}]{\finFiftyInfSixteen};
        \addplot[red!80!black,mark=square,only marks] table[x index={0},y index={1}]{\finFiftyInfThirtytwo};
	    \legend{$A=0$,$A=8$,$A=16$,$A=32$}
	  \end{semilogxaxis}
	\end{tikzpicture}
	\caption{Number of OR nodes inserted and surviving to the solution
    returned by \RLSGP with subtree deletion, using an incomplete training set, stopping once sampled error is at most $A$, $n=50, \Tmax = \infty$, averaged over 500 independent simulations. 
    }
	\label{fig:incomplete-training-ORs}
\end{figure}

Finally, we examine the behaviour of \RLSGP when using an incomplete
training set on larger problem sizes.
The result from Theorem~\ref{thm:incomplete-single-run} relies on the algorithm stopping once a logarithmic sampled error is achieved. We run experiments comparing the performance of \RLSGP when stopping at error 0 or stopping earlier for $n=50$. The average runtimes of the two variants are plotted in Figure~\ref{fig:incomplete-training-results}. The figure confirms our theoretical result that the algorithms generally run in logarithmic time and produce solutions that
contain a logarithmic number of leaf nodes with respect to the training set size. Stopping at 0 error, though, leads to better solutions at the expense of higher runtimes.
Figure~\ref{fig:incomplete-training-ORs} shows the average number of ORs in the final solution. While these are small in number, they grow as the stopping criteria, i.e. the threshold on acceptable sampled error, decreases.

\section{Conclusion} \label{sec:Conclusion}

In this paper, we analysed the behaviour of a variant of the \RLSGP algorithm,
providing rigorous runtime bounds when using the complete truth table to evaluate
solution quality, as well as when using a polynomial sample of possible inputs
chosen uniformly at random. Equipped with a tree size limit and a mutation operator
capable of deleting entire subtrees, \RLSGP is able to efficiently evolve a Boolean 
function -- \andn, the conjunction of $n$ variables -- when given access to both
the binary conjunction and disjunction operators. 

When using the complete truth table to evaluate the quality of solutions, we show
that in expectation, an optimal solution is found within $O(\ell n \log^2 n)$ 
iterations. Experimentally, we see that the GP system is able to find solutions
quicker as $\ell$, the limit on the tree size, increases, suggesting that the
theoretical bound is overly pessimistic in its modelling of the process. Conversely,
solutions with larger tree size limits tend to contain more redundant variables,
suggesting a trade-off between optimisation time and solution complexity.

When sampling a polynomial number of inputs to evaluate program quality,
the evolved solutions are not exactly equivalent to the target function,
but generalise well: any polynomially small
generalisation error can be achieved by sampling a polynomial number of
inputs uniformly at random in each iteration. Our theoretical results predict that
\RLSGP is usually able to avoid inserting ORs in this setting, which is reflected in
our experimental results.

While these results represent a considerable step forward for the theoretical
analysis of GP behaviour, much work remains to be done: apart from the open problem of removing the limit on the tree size,
the analysis could be extended
to cover yet larger function sets (e.g. by also including NOT, allowing the GP to
express any Boolean function), introducing more variables than required by the target
function, or considering a more complex target function where populations and crossover may be required.

\paragraph*{Acknowledgements}  Financial support by the Engineering and
Physical Sciences Research Council (EPSRC Grant No. EP/M004252/1) is gratefully
acknowledged.

\makebibliography{andor}

\end{document}